\pdfoutput=1

\documentclass[11pt]{article}

\usepackage{ACL2023}

\usepackage{times}
\usepackage{latexsym}

\usepackage[T1]{fontenc}

\usepackage[utf8]{inputenc}

\usepackage{microtype}

\usepackage{inconsolata}

\usepackage{paralist}

\usepackage{graphicx}

\usepackage{array}
\usepackage{tabu}
\newcolumntype{M}[1]{>{\centering\arraybackslash}m{#1}}
\newcolumntype{L}[1]{>{\arraybackslash}m{#1}}

\usepackage[table,xcdraw]{xcolor}
\usepackage[normalem]{ulem}
\useunder{\uline}{\ul}{}

\usepackage[acronym]{glossaries}

\usepackage{comment}

\usepackage{subcaption}

\usepackage{url}

\usepackage[inline]{enumitem}

\usepackage{acronym}
\acrodef{BioNLP}[BioNLP]{Biomedical Natural Language Processing}
\acrodef{C-RE}[C-RE]{Concept-level RE}
\acrodef{IAA}[IAA]{Inter-Annotator Agreement}
\acrodef{IE}[IE]{Information Extraction}
\acrodef{KOS}[KOS]{Knowledge Organization System}
\acrodef{LLM}[LLM]{Large Language Model}
\acrodef{M-RE}[M-RE]{Mention-level RE}
\acrodef{NER}[NER]{Named Entity Recognition}
\acrodef{NEL}[NEL]{Named Entity Linking}
\acrodef{NLP}[NLP]{Natural Language Processing}
\acrodef{OIE}[OIE]{Open Information Extraction}
\acrodef{RE}[RE]{Relation Extraction}
\acrodef{SOTA}[SOTA]{State Of The Art}
\acrodef{URI}[URI]{Uniform Resource Identifier}

\title{A Domain-Specific Curated Benchmark for \\ Entity and Document-Level Relation Extraction}

\author{
Marco Martinelli$^{1,*}$ \quad
{\bf Stefano Marchesin}$^{1}$ \quad
{\bf Vanessa Bonato}$^{2}$ \quad
{\bf Giorgio Maria Di Nunzio}$^{1}$ \\
{\bf Nicola Ferro}$^{1}$ \quad
{\bf Ornella Irrera}$^{1}$ \quad
{\bf Laura Menotti}$^{1}$ \quad
{\bf Federica Vezzani}$^{2}$ \quad
{\bf Gianmaria Silvello}$^{1}$ \\
$^{1}$Department of Information Engineering, University of Padova \\
$^{2}$Department of Linguistic and Literary Studies, University of Padova \\
{\small $^{*}$ Corresponding author: \texttt{martinell2@dei.unipd.it}}
} 

\begin{document}
\maketitle
\begin{abstract}
\ac{IE}, encompassing \ac{NER}, \ac{NEL}, and \ac{RE}, is critical for transforming the rapidly growing volume of scientific publications into structured, actionable knowledge. This need is especially evident in fast-evolving biomedical fields such as the gut-brain axis, where research investigates complex interactions between the gut microbiota and brain-related disorders. Existing biomedical \ac{IE} benchmarks, however, are often narrow in scope and rely heavily on distantly supervised or automatically generated annotations, limiting their utility for advancing robust \ac{IE} methods. We introduce \textsc{GutBrainIE}, a benchmark based on more than $1{,}600$ PubMed abstracts, manually annotated by biomedical and terminological experts with fine-grained entities, concept-level links, and relations. While grounded in the gut-brain axis, the benchmark’s rich schema, multiple tasks, and combination of highly curated and weakly supervised data make it broadly applicable to the development and evaluation of biomedical \ac{IE} systems across domains.\footnote{This work was accepted to the Findings of EACL 2026.}

\end{abstract}

\section{Introduction}

Recent studies increasingly associate gut microbiota with neurological and psychiatric disorders like Parkinson’s, Alzheimer’s, Multiple Sclerosis, and mood disorders \cite{carabotti2015gut,ghaisas2016gut,appleton2018gut,cryan2020gut}. PubMed publications on the gut-brain axis more than doubled from 2020 to 2025, increasing from 600 to over 1,500 articles annually. This rapid growth challenges clinicians and researchers to stay updated, identify, and interpret findings in unstructured texts. \ac{NLP} methods for \acf{IE} are crucial in tackling this problem by systematically identifying entities, mapping them to structured knowledge bases, and extracting semantic relations.
\acf{NER} detects and categorizes mentions of biomedical entities such as diseases, chemicals, or anatomical structures. 
\acf{NEL} disambiguates these mentions by linking them to unique identifiers in external knowledge bases, ensuring semantic consistency across texts. 
\acf{RE} identifies and classifies relationships between entities, such as interactions or causal associations. 
Within \ac{RE}, two complementary evaluation settings are commonly distinguished. 
\ac{M-RE} targets specific pairs of mentions in text, making it sensitive to lexical variation and mention boundaries. 
\ac{C-RE} instead operates at the level of linked concepts corresponding to the standard setting adopted by most benchmarks \cite{DETROJA2023200244}.

Despite their importance, implementing these tasks in practice requires substantial amounts of manual annotation, particularly at fine levels of granularity. 
As a result, existing biomedical corpora are often narrow in scope, restricted to a limited set of entity types or relation predicates, confined to sentence-level annotations, and reliant on distant supervision \cite{chen2015study,curationcost,wang2022fine}. 
These limitations hinder the development of \ac{IE} systems capable of handling domains with specialized terminology, diverse concept spaces, and cross-sentence dependencies \cite{tho2024improving,park2024biomedical}. 
The gut-brain axis literature exacerbates these challenges, combining complex terminology, broad conceptual diversity, and long-range relations \cite{liu2021exploring,hong2025dimb}.
To deal with these challenges, we introduce \textsc{GutBrainIE}, a new benchmark for biomedical \ac{IE} based on more than 1,600 PubMed documents annotated by biomedical and terminological experts, trained laypersons, and distantly supervised methods \cite{su2019using}. 
It supports the four complementary \ac{IE} tasks outlined above and is designed both to capture the unique challenges of the gut-brain axis and to serve as a general resource for advancing biomedical \ac{IE}.

The main contributions of this work are: \\
\begin{inparaenum}[(1)]
    \item The first domain-specific corpus focused on the gut-brain axis with a size consistent with established biomedical corpora (cf. Table~\ref{tab:biomedical_ie_datasets}).\\
    \item Manual and automatic annotations based on the most comprehensive and fine-grained schema to date in biomedical \ac{IE}, organized in a stratified structure reflecting different levels of expertise and quality. \\
    \item Standardized benchmark tasks with competitive baselines, evaluation scripts, and leaderboards, facilitating reproducible comparisons across systems.
\end{inparaenum}

The benchmark has been validated through internal experiments with a baseline system and through a large open public evaluation campaign where \textsc{GutBrainIE} served as the reference dataset. 
The dataset is publicly accessible, with training and development sets released with annotations, and the test set provided without ground-truth labels.\footnote{The full benchmark, including code and datasets, is publicly available at: \url{https://zenodo.org/records/16845409}} 
For comparability and reproducibility, each task is hosted on Codabench with official evaluation scripts and leaderboards, allowing system predictions on the test set to be evaluated against the hidden ground truth.\footnote{\url{https://github.com/MMartinelli-hub/GutBrainIE/blob/main/TestData/codabench_competitions.md}}
The rest of the paper is organized as follows: Section~\ref{sec:DataCollection} details data collection and curation. 
Section~\ref{sec:Data Analysis} analyzes the \textsc{GutBrainIE} corpus. 
Section~\ref{sec:Benchmark Settings} introduces the benchmark tasks and discusses its applications. 
Section~\ref{sec:RelatedWork} reviews related corpora. 
Section~\ref{sec:Conclusion} concludes and outlines future work.

\section{Data Collection} \label{sec:DataCollection}

The \textsc{GutBrainIE} benchmark features a large-scale biomedical corpus with manual and automatic annotations for entities, concept-level links, and relations across PubMed documents.
It covers 13 entity types, including both widely used biomedical categories (e.g., \textit{anatomical location}, \textit{bacteria}, \textit{drug}) and entities specific to the gut-brain axis (e.g., \textit{microbiome} and \textit{dietary supplement}).
To account for the frequent occurrence of experimental scenarios in documents, we also introduced specific categories related to medical experiments (e.g., \textit{biomedical technique} and \textit{statistical technique}). 

For what concerns \ac{RE}, \textsc{GutBrainIE} features 17 relation predicates, many of which are overloaded, meaning that the same predicate can link different combinations of entity types depending on context. For example, the predicate \textit{administered} can connect a \textit{chemical}, \textit{drug}, \textit{dietary supplement}, or \textit{food} to either a \textit{human} or \textit{animal} entity. Similarly, the same pair of entity types can be linked through multiple predicates. For instance, a \textit{chemical} can be linked to a \textit{microbiome} entity through either \textit{impact} or \textit{produced by}.  
This many-to-many design originates 55 unique relation triples.
For \ac{NEL}, \textsc{GutBrainIE} links annotated mentions to 6 standardized biomedical vocabularies and a custom-defined ontology for unmatched mentions.


Given the specialized nature of the gut-brain axis, we relied on an in-house team rather than external crowdworkers. This enabled targeted training, regular feedback, and higher annotation consistency. The team comprised 40 trained master’s students in terminography serving as lay annotators and 7 experts, including computer scientists, terminologists specialized in the medical domain, and biomedical specialists with prior experience in evaluation campaigns. 


The manual curation of \textsc{GutBrainIE} followed a four-stage workflow to ensure annotation quality and consistency (illustrated in Figure \ref{fig:annotation_workflow}). 
The preparation involved document retrieval from PubMed and the design of the annotation schema and guidelines. 
The first annotation phase combined \ac{NER} pre-annotations with expert curation. 
The second annotation phase refined \ac{NER} with contributions from both experts and trained lay annotators. 
Finally, the \ac{NEL} phase mapped the set of expert-annotated entity mentions to standardized biomedical vocabularies.

The \textbf{Preparation Phase} began with the retrieval of domain-specific documents from PubMed using two targeted queries, identified by external biomedical experts:
 \texttt{"gut microbiota" AND "Parkinson"} and    \texttt{"gut microbiota" AND "mental health"}.
Two retrieval rounds were conducted on May 9th and October 31st, 2024, resulting in 1,662 documents. After filtering out duplicates and low-relevance documents from earlier publication years (2013-2020), the final collection included 1,647 unique documents.

Following retrieval, we employed an iterative, brainstorming approach for defining the annotation schema and guidelines.
Initially, a representative subset of 100 PubMed abstracts was selected and carefully analyzed by a focus group of expert annotators in collaboration with biomedical domain experts and terminologists, leading to the identification of a core set of domain-specific definitions, based on which we drafted a first annotation schema, defining the entities and relations of interest.
This schema was further refined by extending the analysis to the full set of retrieved documents, resulting in a finalized structure comprising 13 entity types and 17 fine-grained relation predicates (see Figure \ref{fig:conceptual_schema} and Tables \ref{tab:entity_definitions}-\ref{tab:relation_definitions} in the appendix). 

After these stages, the expert annotators' team defined a detailed set of annotation guidelines, with the final goal of obtaining high-quality annotations that are consistent through different annotators and documents.
Inspired by prior works such as BioRED \cite{biored}, BC5CDR \cite{bvcdr}, and BioASQ-QA \cite{bioasq-qa}, these guidelines detail the end-to-end annotation process to be followed for each document, including labeling rules, edge case handling, and examples of typical annotations and mistakes.\footnote{The annotation guidelines are available at: \url{https://zenodo.org/records/16845409/files/GutBrainIE_2025_Annotation_Guidelines.pdf}}  
Before starting the manual annotation phases, a hands-on training session was conducted with all expert annotators, during which they jointly annotated a set of abstracts for both entity mentions and relations while reviewing and refining the guidelines to ensure consistent interpretation and resolve any ambiguities early on. Two additional sessions were conducted with biomedical experts, following the same process and providing domain-specific feedback for further guidelines adjustments.

\textbf{First Annotation Phase.}
Before starting the actual curation, we adopted an automatic annotation support strategy for entity mentions to reduce the manual effort required from annotators and accelerate the annotation process \cite{ganchev2007semi,greinacher2018dalphi,mikulova2023quality}.
We began by selecting a representative sample of ten documents and asked two terminologists to manually annotate all entity mentions. 
We then selected a competitive zero-shot \ac{NER} model -- GLiNER \cite{zaratiana-etal-2024-gliner} -- and systematically experimented with different pre-trained checkpoints, temperature values, and post-processing strategies, comparing the model's predicted mentions with the manually annotated ones.
Based on these experiments, we selected the \textit{NuNER Zero} checkpoint and adopted a temperature value of 0.8 to obtain a higher recall and minimize the number of entity mentions annotated by terminologists but missed by the model \cite{nuner}. 
To preliminarily assess the impact of these pre-annotations, we asked the same terminologists to annotate ten additional documents using the GLiNER-generated pre-annotations. Without pre-annotations, they needed around 30 minutes per document to fully annotate its entity mentions. With pre-annotations, the time dropped to 10-15 minutes, depending on the document's complexity. 

All pre-annotated documents were annotated with \textit{MetaTron}, a free and publicly available web-based annotation platform \cite{metatron}.
In this first annotation phase, only expert annotators were involved. Each of them received 20 unique documents along with a shared set of 5 ``honeypot documents''. 
The latter were introduced to compute \ac{IAA}, enabling direct comparison of annotations across annotators and the assessment of their agreement and consistency.
To avoid bias or information leakage, we instructed annotators not to consult with each other until all their annotations were completed \cite{omarbook}. 

In computing \ac{IAA}, two annotations agree only if their text spans, labels (and predicates, in the case of relations) exactly matched.
Using Fleiss’ $\kappa$ \cite{fleiss1971measuring} and Mean Pairwise Cohen’s $\kappa$ \cite{cohen1960coefficient} as metrics, we observed strong agreement for \ac{NER} (0.89 and 0.88, respectively), aligned with previous biomedical datasets such as BioRED \cite{biored} and NLM-Gene \cite{islamaj2021nlm}. 
Agreement on \ac{RE} was lower (0.43 for both metrics) but consistent with prior works involving complex relation annotation, reflecting the semantic difficulty of the gut-brain axis domain \cite{kim2013gro}.

To complement the \ac{IAA} analysis, a subgroup of two expert annotators, selected for domain expertise and prior annotation experience, manually revised all the documents annotated during this phase. 
This step led to the correction or removal of 204 entities and 135 relations. For entities, roughly one-third of edits were due to incorrect text spans, another third to mislabeling, and the rest to overly generic mentions that violated annotation guidelines. 
For relations, about one-third were revised for incorrect directionality, another third for wrong predicate assignment, and the remainder for insufficient textual support. 
This review also allowed for the identification of issues in the annotation guidelines, which were refined accordingly. All annotations were then retroactively updated to align with the updated guidelines. 

\textbf{Second Annotation Phase.}
Prior to the second phase, we fine-tuned GLiNER on the final revised annotations from the first phase to improve the quality of pre-annotations. 
In this second phase, each expert annotator labeled 40 new documents, while layperson annotators were introduced and assigned 24 documents each, divided into four batches. In each batch, one honeypot document previously annotated by experts was randomly inserted. Laypeople were required to annotate a minimum of one full batch, ensuring that each of them was labeling at least one honeypot document. 

To assess the quality of layperson annotations, we computed Cohen's $\kappa$ on their annotated honeypot documents against the reference version curated by experts \cite{cohen1960coefficient}.
Agreement for \ac{NER} was moderate and acceptable (0.50), indicating that with adequate training and supervision non-expert annotators can contribute reliable entity annotations. 
In contrast, \ac{RE} agreement was low (0.17), largely due to the annotation of relations against the guidelines. 

At the end of this phase, expert annotators conducted a final review meeting to evaluate the second batch of annotated documents, addressing any unresolved issues and making corrections to ensure consistency across the full collection.

\textbf{Named Entity Linking Phase.}
We performed entity linking on expert annotations only, as laypeople and automatic mentions often lack the precision required for reliable concept mapping.

We applied normalization based on heuristics derived from manual analysis of expert-curated annotations. This involved removing HTML tags and special symbols, using regular-expression substitutions to handle terminological variants and expand common acronyms into their full forms. 
Then, we implemented a three-stage linking approach.
For each annotated mention at first we attempted exact string matching against the reference biomedical vocabularies defined for the entity's label. Resources were queried sequentially in 
a predefined priority order for each entity type, aiming to prioritize linkages to the largest and most established biomedical vocabularies. 
If no exact match was found, we computed the embeddings with BiomedBERT and then calculated the cosine similarity between the mention and all possible candidates in the reference vocabularies \cite{pubmedbert}. 
Only candidates with similarity scores above a dynamic cut-off were retained, computed using the central limit theorem to approximate the distribution of similarity scores as normal and ensure statistical reliability \cite{kwak2017central}.
If both previous steps failed, we assumed the mention was not in the reference vocabularies. In such cases, we first attempted to map the mention to an existing entry in our custom ontology. If no match was found, we added a new concept to our ontology by prompting an \acs{LLM} (\texttt{LLaMA3-8B-8192} in this case) with the sentence containing the mention (highlighted with inline markers) to generate a pertinent definition (see Figure \ref{fig:llm_prompt_example} for an example). 
In all stages, if multiple candidate links were returned, we manually selected the most appropriate one.
Finally, all generated linkages and individuals created in our custom ontology were manually reviewed and verified by experts to ensure consistency and resolve ambiguous mappings.  
The accuracy of the \ac{NEL} annotations was estimated at $0.915 \pm 0.0473$ using a sampling-based evaluation framework with statistical guarantees \cite{martinelli2026efficientreliableestimationnamed}.

\textbf{Automatically-Annotated Data Collection.}
After finalizing the manual annotations, we automatically annotated the remaining unlabelled documents from the original retrieval. 
For \ac{NER}, we fine-tuned again the GLiNER model using the full set of expert-annotated data. For \ac{RE}, we introduced ATLOP, a model leveraging adaptive thresholding and localized context pooling to effectively capture long-tail relations \cite{atlop}.
This feature is critical for the \textsc{GutBrainIE} corpus, where various low-frequency relations play a significant role in accurately representing biomedical interactions.
ATLOP was trained with expert annotations as primary supervision and student annotations as weak supervision.
Although no manual revision was performed on this data, we applied a post-processing step to ensure adherence to the same schema and guidelines as the human-curated data.

\textbf{Collection Overview.}
Once all the documents from the original retrieval were annotated, either manually or automatically, we organized them into four quality-based folds reflecting the reliability of the annotations and the level of expertise of the annotators involved:
\begin{enumerate*}
    \item \textbf{Platinum}: highest-quality annotations produced by experts during the first manual annotation phase and internally reviewed by a dedicated subgroup. 
    \item \textbf{Gold}: high-quality annotations created by experts during the second manual annotation phase. 
    \item \textbf{Silver}: annotations of intermediate quality produced by trained layperson annotators under expert supervision.
    \item \textbf{Bronze}: automatically generated annotations. 
\end{enumerate*}

The training set includes all four quality folds. As for the Development and Test sets, they contain only expert annotations (Platinum and Gold). 

Each annotation is tagged with an anonymized annotator ID, with expert annotators identified as \texttt{expert[1–7]} and automatic annotations as \texttt{automatic}.
For layperson annotators, we clustered them into two groups (A and B) based on annotation quality: Group A achieved the highest overlap with experts ($\geq65\%$ for entities and $\geq40\%$ for relations), whereas Group B showed lower agreement.
Accordingly, student annotators are identified as \texttt{student[A/B]}, depending on the reliability cluster to which they have been assigned.

This tiered annotation quality, along with annotator metadata, allows models to be trained or tuned by weighting or filtering annotations differently based on their reliability, supporting training strategies such as reliability-aware loss weighting \cite{lin2019reliability,ibrahim2020confidence,guo2024improving} and selective denoising 
\cite{ghosh2023aclm,alomar2023data}. 
Table \ref{tab:gutbrainie_dataset_summary} shows the distribution of documents and annotations per fold.

\begin{table}[t] 
\centering
\caption{\textsc{GutBrainIE} dataset statistics. }
\label{tab:gutbrainie_dataset_summary}
\resizebox{0.45\textwidth}{!}{%
\begin{tabular}{|l|r|r|r|r|r|}
\hline
\textbf{Collection}                                                & \textbf{No. Docs} & \textbf{No. Ents} & \textbf{Ents/Doc} & \textbf{No. Rels} & \textbf{Rels/Doc} \\ \hline
Silver         & 499   & 15,275 & 30.61 & 10,616 & 21.27 \\
Gold           & 208   & 5,192  & 24.96 & 1,994  & 9.59  \\
Platinum       & 111   & 3,638  & 32.77 & 1,455  & 13.11 \\
Dev            & 40    & 1,117  & 27.93 & 623    & 15.58 \\
Test           & 40    & 1,237  & 30.92 & 777    & 19.42 \\ \hline
Manual & 898   & 26,459 & 29.46 & 15,465 & 17.22 \\
\begin{tabular}[c]{@{}l@{}}Automatic \\ (Bronze)\end{tabular} & 749               & 21,420            & 28.51             & 8,533             & 10.90             \\ \hline
Overall        & 1,647 & 47,879 & 29.03 & 23,998 & 14.35 \\ \hline
\end{tabular}%
}
\end{table}

\section{Data Analysis} \label{sec:Data Analysis}
To better contextualize our corpus features and strengths, we compare it against a selection of widely used general-purpose and biomedical \ac{IE} datasets.
We excluded most distantly supervised datasets and focused on manually annotated ones that provide reliable labels and are directly comparable to ours \cite{amin2020data, amin2022meddistant19}.

\textbf{Data Size.}
The \textsc{GutBrainIE} corpus consists of 1,647 documents, of which 898 are manually annotated. This positions it among the largest manually curated biomedical corpora, with a size comparable to BioRED (600 documents) and NCBI disease (793). 
In terms of length, \textsc{GutBrainIE} documents average 235 words, which is aligned with widely used biomedical and general-purpose datasets such as BioRED (240 words), JNLPBA (240), NCBI Disease (227), and DocRED (210) 
\cite{biored, collier2004introduction, dougan2014ncbi, docred}.
Table \ref{tab:biomedical_ie_datasets} reports overall statistics for \textsc{GutBrainIE} and several representative \ac{IE} datasets. We report aggregated statistics for both the full collection and the manually annotated subset.

\begin{table*}[!htb]
\centering
\caption{Comparison of \textsc{GutBrainIE} with the main \ac{IE} datasets. Reported annotation counts include both entities and relations for datasets featuring \ac{RE}. and the ``\textit{Task(s)},, column indicates the tasks supported by each dataset. Rows highlighted in red correspond to biomedical \ac{IE} datasets. while those in green indicate general-purpose \ac{IE} datasets.}
\label{tab:biomedical_ie_datasets}
\resizebox{\textwidth}{!}{%
\begin{tabular}{l l l c c c c c c c}
\hline
Dataset &
  Year &
  Task(s) &
   &
  No. Docs &
  No. Entity Mentions &
  No. Entity Types &
  No. Relations &
  No. Relation Predicates. &
  No. Annotations \\ \hline
  \rowcolor[HTML]{ECF4FF} 
\textsc{GutBrainIE} (Manual) &
  2025 &
  NER. NEL. RE &
   &
  898 &
  26,459 (29.46 per doc) &
  13 &
  15,465 (17.22 per doc) &
  17 &
  41,924 (46.69 per doc) \\
  \rowcolor[HTML]{ECF4FF} 
\textsc{GutBrainIE} (Manual+Automatic) &
  2025 &
  NER. NEL. RE &
   &
  1,647 &
  47,879 (29.07 per doc) &
  13 &
  23,998 (14.57 per doc) &
  17 &
  71,877 (43.64 per doc) \\ \hline
  \rowcolor[HTML]{FFECEC} 
JNLPBA \cite{collier2004introduction}      & 2004 & NER          &  & 2404  & 59,963 (24.94 per doc) & 5 & --     & -- & 59,963 (24.94 per doc) \\
  \rowcolor[HTML]{FFECEC} 
EU-ADR \cite{van2012eu}      & 2012 & NER. RE      &  & 300   & 7,011 (23.37 per doc)  & 3 & 2,436 (8.12 per doc)  & 3  & 9,447 (31.49 per doc)  \\
  \rowcolor[HTML]{FFECEC} 
BioNLP-CG \cite{pyysalo2013overview}  & 2013 & NER. RE      &  & 600   & 21,683 (36.14 per doc) & 4 & 917 (1.53 per doc)   & 4  & 22,600 (37.67 per doc) \\
  \rowcolor[HTML]{FFECEC} 
NCBI Disease \cite{dougan2014ncbi} & 2014 & NER. NEL     &  & 793   & 6,892 (8.69 per doc)   & 1 & --     & -- & 6,892 (8.69 per doc)   \\
  \rowcolor[HTML]{FFECEC} 
CHEMDNER \cite{krallinger2015chemdner}    & 2015 & NER          &  & 10,000 & 84,355 (8.44 per doc)  & 1 & --     & -- & 84,355 (8.44 per doc)  \\
  \rowcolor[HTML]{FFECEC} 
BC5CDR \cite{bvcdr}      & 2016 & NER. NEL. RE &  & 1500  & 12,850 (8.57 per doc)  & 2 & 3,116 (2.08 per doc)  & 2  & 15,966 (10.64 per doc) \\
  \rowcolor[HTML]{FFECEC} 
NLM-Gene \cite{islamaj2021nlm}    & 2021 & NER. NEL     &  & 550   & 15,553 (28.28 per doc) & 1 & --     & -- & 15,553 (28.28 per doc) \\
  \rowcolor[HTML]{FFECEC} 
BioRED \cite{biored}      & 2022 & NER. NEL. RE &  & 600   & 20,419 (34.03 per doc) & 6 & 6,503 (10.84 per doc) & 8  & 26,922 (44.87 per doc) \\ \hline
  \rowcolor[HTML]{EFFFEC} 
DocRED (Manual) \cite{docred} &
  2019 &
  NER. NEL. RE &
   &
  5,053 &
  98,533 (19.50 per doc) &
  6 &
  56,798 (11.24 per doc) &
  96 &
  155,331 (30.74 per doc) \\
  \rowcolor[HTML]{EFFFEC} 
HacRED \cite{hacred} &
  2021 &
  NER. NEL. RE &
   &
  9,231 &
  98,772 (10.70 per doc) &
  10 &
  56,798 (6.15 per doc) &
  26 &
  155,570 (16.85 per doc) \\ 
  \rowcolor[HTML]{EFFFEC} 
  Re-DocRED \cite{tan2022revisiting} &
  2022 &
  NER. NEL. RE &
   &
  4,053 &
  78,628 (19.40 per doc) &
  6 &
  120,664 (29.77 per doc) &
  96 &
  199,292 (49.17 per doc) \\ \hline
\end{tabular}%
}
\end{table*}

\textbf{Annotated Entities and Relations.}
\textsc{GutBrainIE} contains an average of 29 entity mentions and 15 relations annotated per document, comparable to BioRED (34 mentions, 11 relations) and exceeding other general-purpose corpora like DocRED (26 mentions, 11 relations) and HacRED (11 mentions, 6 relations), as well as biomedical datasets such as JNLPBA (25 mentions, 15 relations), NCBI Disease (9 mentions), and CHEMDNER (8 mentions) \cite{biored,hacred,collier2004introduction,dougan2014ncbi,krallinger2015chemdner}.
Nevertheless, while the average number of entity mentions and relations per document in \textsc{GutBrainIE} is comparable to other widely used corpora, our dataset offers greater granularity, more than doubling the number of entity and relation types compared to BioRED (7 and 8, respectively), which is the most fine-grained manually curated biomedical corpus to date \cite{biored}.

\textbf{Linkage Statistics.}
As stated in Section \ref{sec:DataCollection}, all entity mentions in the Gold and Platinum folds have been linked to standardized medical vocabularies, resulting in a total of 1,819 unique \acsp{URI} assigned across 11,184 annotated mentions.

Each entity type is linked, on average, to concepts from four different vocabularies, with preference given to authoritative resources in the field. In particular, we prioritized mappings to UMLS, a widely used metathesaurus, to maximize interoperability with other biomedical datasets \cite{umls}. Reference vocabularies associated with each entity type are reported in Table \ref{tab:links_definitions}.

\section{Benchmark Setup and Validation} \label{sec:Benchmark Settings}
The \textsc{GutBrainIE} benchmark natively supports four tasks: \ac{NER}, \ac{NEL}, \ac{M-RE}, and \ac{C-RE}. 

The \ac{NER} task requires identifying and classifying entity mentions into 13 predefined entity types. Each mention is represented as a tuple including: location (title or abstract), character offsets (start and end positions), text span, and entity label.
A prediction is considered correct only if it exactly matches a ground truth entry in all these fields.

The \ac{NEL} task extends \ac{NER} by additionally requiring the linkage to a \acs{URI} from one of the reference vocabularies.
Here, predicted entities must match ground truth annotations in all \ac{NER} fields plus \acs{URI} to be considered correct. 

The \ac{M-RE} task involves detecting and classifying relations between pairs of entity mentions. Each relation is expressed as a 5-tuple:
\texttt{(subject text span, subject label, relation predicate, object text span, object label)}.
Predicted relations are considered correct only if they fully match a ground truth tuple.

The \ac{C-RE} task mirrors \ac{M-RE} but evaluates predictions at the concept level, with relations represented as 5-tuples:
\texttt{(subject URI, subject label, predicate, object URI, object label)}.
For example, a predicted relation involving ``\textit{Parkinson}'' would match a ground truth relation comprising ``\textit{Parkinson’s disease}'' if both mentions are linked to the same \acs{URI}.
The correctness of predicted relations is assessed as in \ac{M-RE}.

\textbf{Benchmark Validation.} \label{sec:Experiments}
To assess the complexity and utility of the \textsc{GutBrainIE} benchmark, we conducted internal experiments across all four proposed benchmark tasks, adapting the automatic annotation system to serve as a baseline system. Moreover, \textsc{GutBrainIE} has been externally validated by featuring the \ac{NER} and \ac{M-RE} tasks in an international evaluation campaign, which attracted 17 participating teams.   
Internal and external results have been analyzed to assess performance variations across tasks and their relative difficulty, and to validate the intended progression in complexity. 
Finally, we estimated human performance by comparing non-expert annotations on shared honeypot documents with our baseline system.

All tasks are evaluated using standard \ac{IE} metrics of Precision ($P$), Recall ($R$), and F1-score ($F_1$), with both macro- and micro-averaging. 
Let $TP_{\ell}$, $FP_{\ell}$, and $FN_{\ell}$ denote the number of True Positives, False Positives, and False Negatives for label $\ell$. The label set $\mathcal{L}$ is defined as the set of 13 entity types for the \ac{NER} and \ac{NEL} tasks, and the 55 possible relation triples \texttt{(subject label, predicate, object label)} for the \ac{M-RE} and \ac{C-RE} tasks.
Before evaluation, duplicate predictions are removed and overlapping ones are merged, keeping the one with the longest text span.
Micro-averaged $F_{1}$ is used as the reference metric for all tasks, as it better captures the balanced effectiveness of \ac{IE} systems and accounts for class imbalance.

\textbf{Baseline System.} \label{sec:BaselineSystem}
To provide reference performances, we derived a baseline system from the one used for pre-annotating documents. 
To perform \ac{NER} we adopted GLiNER fine-tuned from the \texttt{NuNERZero} checkpoint on the Platinum, Gold, and Silver collections \cite{zaratiana-etal-2024-gliner, nuner}. At inference time, we applied a 0.6 confidence threshold and merged adjacent and overlapping spans.
For \ac{RE}, we used ATLOP, giving the document text and the entities predicted by GLiNER as inputs \cite{atlop}. 
ATLOP is trained using the Platinum, Gold, and Silver collections as primary supervision, while the Bronze collection is used as distantly supervised data.
Inferred relations are processed to retain only triples defined in the annotation schema.
The \ac{NEL} module links GLiNER-predicted entities by applying the same three-stage hierarchical approach described above. 
Baseline results are reported in Table \ref{tab:baseline_results}. 

\begin{table}[t]
\centering
\caption{Baseline results across all four \textsc{GutBrainIE} supported tasks.}
\label{tab:baseline_results}
\resizebox{0.4\textwidth}{!}{%
\begin{tabular}{l|ccc|ccc}
\hline
\multicolumn{1}{c|}{} & \multicolumn{3}{c|}{Macro-avg} & \multicolumn{3}{c}{Micro-avg} \\ 
Task                  & $P$        & $R$        & $F_1$       & $P$        & $R$        & $F_1$      \\ \hline
NER                   & 0.69     & 0.71     & 0.70     & 0.76     & 0.82     & 0.79    \\
NEL                   & 0.40     & 0.41       & 0.39       & 0.50       & 0.54       & 0.52      \\
M-RE                  & 0.35     & 0.18     & 0.21     & 0.50     & 0.25     & 0.33    \\
C-RE                  & 0.30       & 0.18       & 0.20       & 0.38       & 0.20       & 0.27      \\ \hline
\end{tabular}%
}
\end{table}

\textbf{External Validation.}
To exhaustively validate the \textsc{GutBrainIE} benchmark, the \ac{NER} and \ac{M-RE} tasks have been offered within an international evaluation campaign.
Participants were provided with the dataset, annotation guidelines, and the baseline system. Teams were free to employ any model architecture, training strategy, or external resource.

A total of 17 distinct teams participated: 15 in \ac{NER} and 12 in \ac{M-RE}, submitting 101 and 95 runs respectively. 
Considering micro-averaged $F_{1}$ as the reference metric, 8 teams outperformed the baseline on \ac{NER} (38 runs) and 5 on \ac{M-RE} (24 runs). 
These results highlight both the competitiveness of the baseline system and the clear margin for improvement (see Tables \ref{tab:NER_leaderboard} and \ref{tab:TM-RE_leaderboard} in the annex for team scores). 
For full details on approaches and results, see \cite{martinelli2025overview}.


\textbf{Model Performance.}
The results obtained by the baseline and the teams participating in the campaign provide valuable insights into the strengths and limitations of current \ac{NLP} methods across biomedical \ac{IE} tasks of increasing complexity.

For \ac{NER}, most teams relied on supervised fine-tuning of pre-trained biomedical transformers, including PubMedBERT, BioBERT, BioLinkBERT, and BioMedELECTRA \cite{pubmedbert,biobert,linkbert,electra}. 
Several participants also adopted GLiNER, the same architecture used in the baseline, but with different checkpoints and fine-tuning strategies \cite{zaratiana-etal-2024-gliner}. 
Many submissions further improved performance through ensemble methods, either by combining different models or multiple instances of the same model trained with different configurations, seeds, or data splits. 
Indeed, while most systems were trained exclusively on the manually curated Platinum, Gold, and Silver collections, a few teams also included the Bronze collection, using reweighting or filtering strategies to mitigate its noise. However, these attempts led to inconsistent improvements, suggesting that the benefits of using automatically annotated data in high-granularity \ac{IE} tasks remain limited.

\ac{M-RE} results reflected the inherent high complexity of this task. 
Most teams approached \ac{M-RE} as a supervised classification problem over entity mention pairs predicted by an upstream \ac{NER} module. To address class imbalances and long-tail relations, participants employed negative sampling, class-weighted loss functions, and filtering heuristics, and a few explored more advanced architectures such as query-based encoders and hypergraph models \cite{feng2019hypergraph}.

Nevertheless, the baseline system achieved competitive results, ranking close to the median across both subtasks, indicating the difficulty of achieving substantial improvements on our benchmark.
In contrast, prompt-based and zero-shot \acs{LLM} approaches used by a few teams to perform end-to-end \ac{NER} and \ac{RE} performed substantially worse, indicating the current limitations of \acsp{LLM} for domain-specific and fine-grained \ac{IE} tasks.

Overall, these results demonstrate that \ac{IE} systems face numerous challenges in achieving robust performance on the \textsc{GutBrainIE} benchmark, as participants who achieved substantial improvements over the baseline did so by adopting sophisticated architectures, advanced training strategies, and computationally intensive solutions. 
This underscores that our benchmark offers ample opportunities for methodological research and innovation.

\textbf{Human Performance.}
To estimate human-level performance on the benchmark tasks proposed in the evaluation campaign, we evaluated layperson annotations on the shared honeypot documents.
Each set of annotations by a student was treated as an individual system submission and evaluated using the same script and metrics applied to participant test runs, with the final annotated version of each honeypot document used as ground truth.
To establish a fair comparison, we re-trained our baseline system leaving out the honeypot documents to prevent data leakage, then ran inference on the honeypot set and evaluated its predictions.

For the \ac{NER} task, all laypeople achieved Precision ($P$), Recall ($R$), and $F_1$ scores above 0.40, with average scores of 0.79 $P$, 0.77 $R$, and 0.77 $F_1$. 
Although lower, results were still robust for \ac{M-RE}, where, on average, laypeople scored higher on $P$ (0.61) and slightly lower on $R$ (0.52) and consequently on $F_1$ (0.53). 
The baseline system achieved a micro-averaged $P$, $R$, and $F_1$ of 0.83 for \ac{NER}, and 0.44 $P$, 0.31 $R$, and 0.37 $F_1$ for \ac{M-RE}.
These results indicate that, while \ac{NER} can be effectively tackled by automatic systems with performance comparable to non-expert annotators, \ac{RE} remains significantly more complex. In this task, layperson annotators consistently outperformed the baseline across all metrics, highlighting the semantic and contextual difficulty of our benchmark.

\textbf{Discussion.}
The experimental results confirm that \textsc{GutBrainIE} is a robust and well-designed benchmark, presenting significant challenges for current \ac{IE} systems, especially in \ac{RE} tasks, which demand methodological advancements and refined \ac{IE} approaches. 
External validation shows that established Bio\ac{NLP} methods are highly effective across tasks. In contrast, emerging methods based on \acsp{LLM} are still immature and do not match the performance of supervised systems in specialized domains, indicating that their potential remains largely unexploited for fine-grained \ac{IE}.
Moreover, we found out that models trained on smaller subsets of expert-annotated data consistently outperformed those trained on larger datasets that included the automatically annotated noisier portion of the corpus, reinforcing the importance of high-quality annotations in \ac{IE} and validating the tiered quality structure of the \textsc{GutBrainIE} corpus.

\section{Related Work} \label{sec:RelatedWork}
A variety of datasets have been developed to support \ac{IE} in the biomedical domain, particularly for \ac{NER}, \ac{NEL}, and \ac{RE}. Early biomedical corpora such as JNLPBA \cite{collier2004introduction}, EU-ADR \cite{van2012eu}, BioNLP-CG \cite{pyysalo2013overview}, and BC5CDR \cite{bvcdr} focused on a small number of entity types, including genes, diseases, and chemicals, and helped establish benchmarks for single-type \ac{NER} systems. However, systems trained on one or a few entity types often fail to generalize across broader arrays of biomedical concepts. Moreover, various studies demonstrated that in Bio\ac{NLP} multi-type \ac{NER} models can perform comparably or better than single-type models, showing greater capabilities in leveraging contextual information and handling ambiguities \cite{crichton2017neural,wang2019cross}.
For what concerns \ac{RE}, the high cost of manual annotations led most biomedical \ac{IE} datasets to rely on distant supervision for inferring relations, inevitably introducing noise and incorrectness \cite{curationcost,amin2020data,amin2022meddistant19}.
Recognizing that gap, BioRED introduced a fine-grained dataset covering six biomedical entity types and eight relation predicates, with manually curated \ac{NER}, \ac{NEL}, and document-level \ac{RE} annotations \cite{biored}. 


Compared to existing resources, \textsc{GutBrainIE} introduces a large domain-specific \ac{IE} benchmark with manual annotations for entities, concept-level linkages, and relations divided into a multi-tiered quality structure. 
To our knowledge, \textsc{GutBrainIE} provides the most fine-grained annotation schema to date for both entities and relations in the biomedical domain. 

\section{Conclusions and Future Work} \label{sec:Conclusion}
In this work we presented \textsc{GutBrainIE}, a comprehensive \ac{IE} benchmark focusing on the emerging biomedical research area of the gut-brain axis. 
\textsc{GutBrainIE} provides a large domain-specific dataset manually curated, supporting four well-defined tasks of increasing complexity (\ac{NER}, \ac{NEL}, \ac{M-RE}, and \ac{C-RE}), each accompanied by standardized evaluation measures and a competitive baseline system.
To demonstrate its impact and practical utility, we featured two of its tasks (\ac{NER} and \ac{M-RE}) as part of an international evaluation campaign, which attracted 17 participating teams with nearly 200 system submissions. 
Experimental results indicate that current \ac{NER} systems are effective even in specialized domains, while \ac{RE} remains a significantly more challenging task in domains requiring deep contextual and semantic understanding. 
\textsc{GutBrainIE} has been developed to support \ac{IE} research at large, proposing a reliable and challenging evaluation framework for settings characterized by domain specificity, limited training data, and complex terminology.
In future work, we plan to extend the \textsc{GutBrainIE} corpus by annotating documents related to additional neurodegenerative diseases (e.g., Alzheimer’s, Multiple Sclerosis) and to further enhance the quality of existing data by manually revising the Silver and Bronze folds.


\section*{Limitations}
While \textsc{GutBrainIE} is a novel and high-quality benchmark, it has a few limitations.
First, from the analysis of layperson annotations conducted to cluster them into the two reliability groups (cf. Section \ref{sec:DataCollection}), we observed that the Silver collection includes annotations that are not fully consistent with those in the Platinum and Gold collections.
Moreover, the automatically generated annotations of the Bronze collection exhibit notable noise. As observed in experiments, incorporating these lower-quality folds directly into training might degrade model performance.

Concerning the annotation workflow, it was conducted in separate batches, which may have introduced inconsistencies in annotations. Future annotation cycles could leverage active learning techniques with continuous model-in-the-loop feedback to enhance consistency and reduce manual effort.

Finally, given that the involvement of biomedical domain experts played a critical role in the development of the conceptual schema and annotation guidelines, future annotation cycles could further benefit from their integration as external reviewers or adjudicators to enhance annotation quality and accuracy.

\section*{Ethical Considerations}
Below we detail relevant ethical aspects related to the construction and dissemination of the \textsc{GutBrainIE} dataset:
\begin{itemize}
    \item \textbf{Intellectual property and data sources}: the \textsc{GutBrainIE} dataset comprises only titles and abstracts of biomedical articles retrieved from PubMed, a publicly accessible electronic database. These documents are available for reuse under terms that permit research and educational purposes, and no full-text content or content under restricted licenses has been included.

    \item \textbf{Annotators privacy}: although we collected information about the annotators to manage task assignments and quality control during the annotation process, the released version of the dataset includes only anonymized identifiers, designed to preserve the utility of the data while ensuring that no individual annotator can be personally identified from the released dataset.

    \item \textbf{Annotators compensation}: all annotations were performed by volunteers, who were informed in advance that no monetary compensation would be provided. Participation was entirely voluntary and conducted in a non-commercial academic context.

    \item \textbf{Data transparency and characteristics}: detailed information about the annotation schema, workflow, and data characteristics is provided in the appendix and in the annotation guidelines file.

    \item \textbf{Potential data quality issues}: although a significant portion of data has been manually annotated by experts following strict guidelines, we acknowledge the possibility of residual noise or inconsistencies also in the highest-quality Gold and Platinum collections. Such issues are common in most publicly available corpora and are not expected to critically impact downstream applications.

    \item \textbf{Use of generative AI}: during the preparation of this work, the authors used GPT-4o and Grammarly for grammar and spelling checks. After using these tools, the authors reviewed and edited the content as needed and takes full responsibility for the publication’s content.
\end{itemize}

We believe that the publication and use of \textsc{GutBrainIE} will contribute positively to the development of robust and effective \ac{IE} systems across a variety of semantically rich and complex domains, including but not limited to \ac{BioNLP} and health-related applications.

\section*{Acknowledgements}
This project has received funding from the HEREDITARY Project, as part of the European Union's Horizon Europe research and innovation programme under grant agreement No GA 101137074.     

\bibliography{anthology,custom}
\bibliographystyle{acl_natbib}

\clearpage
\newpage
\appendix
\section{\textsc{GutBrainIE} Benchmark at a Glance}
\begin{figure*}[!htb]
    \centering
    \includegraphics[width=0.85\linewidth]{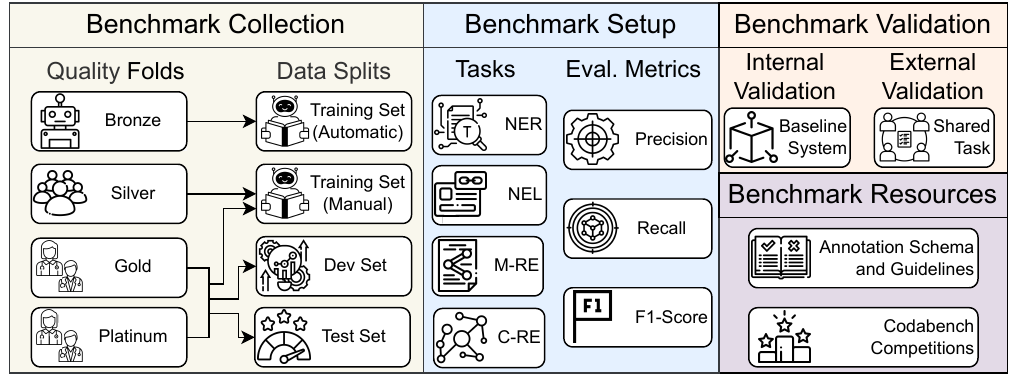}
    \caption{Summary of the main features and contributions of the \textsc{GutBrainIE} benchmark.}
    \label{fig:benchmark_summary}
\end{figure*}

\begin{figure*}[!htb]
    \centering
    \includegraphics[width=\linewidth]{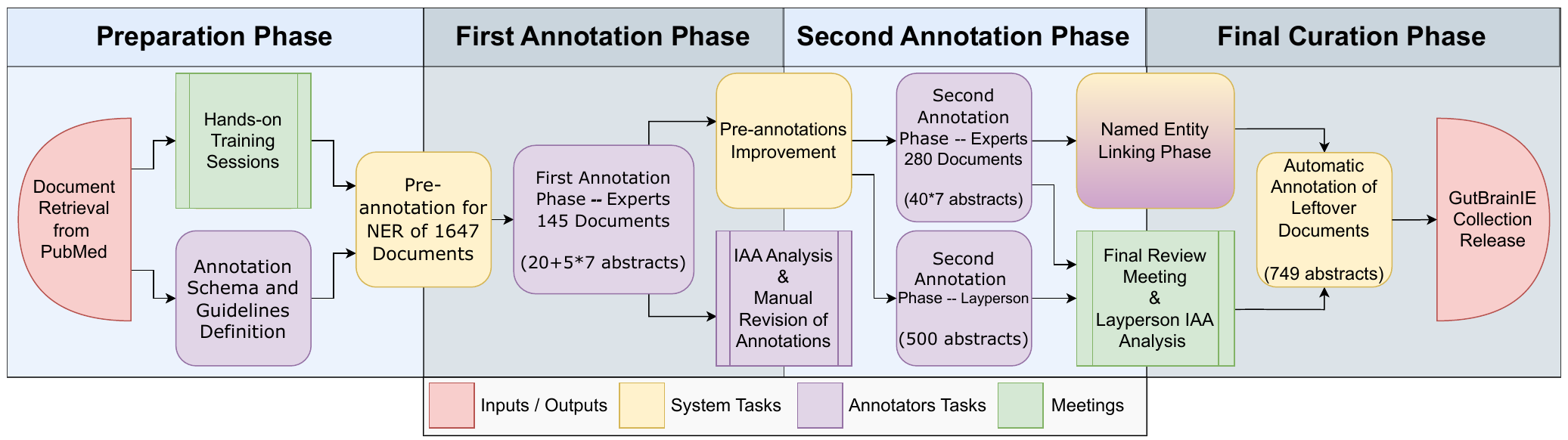}
    \caption{The four-stage workflow followed in the curation of the \textsc{GutBrainIE} collection.}
    \label{fig:annotation_workflow}
\end{figure*}

The entire \textsc{GutBrainIE} benchmark is summarized in Figures \ref{fig:benchmark_summary}-\ref{fig:annotation_workflow}.
In particular, Figure \ref{fig:annotation_workflow} shows the four-stage annotation workflow adopted to build the \textsc{GutBrainIE} collection. The \textit{Named Entity Linking Phase} is highlighted with a purple-to-yellow gradient to indicate that it combines automatic and manual annotation. Automatic annotations for the remaining documents were produced using the system from which our baseline system is derived.

\section{Shared Task Results}
The results obtained by teams participating in the international shared task are summarized in Tables \ref{tab:NER_leaderboard} (\ac{NER}) and \ref{tab:TM-RE_leaderboard} (\ac{M-RE}) \cite{nentidis2025overview,martinelli2025overview}. For each participating team, we report the performance of their best submitted run, considering the micro-averaged $F_1$-score as the reference metric. In both tables, the entry \textsc{BASELINE} (highlighted in blue) corresponds to the results achieved by our baseline system.

Across submissions on both tasks, participants largely relied on supervised pipelines built around biomedical transformer encoders, with performance gains mainly coming from system-level modifications rather than fundamentally different paradigms. For \ac{NER}, strong systems consistently relied on fine-tuned transformers used for token classification (often enhanced with structured decoding such as CRF-style label dependency modeling). Performance gains have been observed with ensembling, confidence thresholding, and lightweight post-processing to correct boundaries and merge overlaps. Several teams experimented with incorporating lower-quality (bronze/silver) annotations or pseudo-labeled data, typically after cleaning or reweighting, suggesting that additional weak supervision can be beneficial in some configurations but requires careful filtering to avoid introducing noise. In a minority of cases, generation- or schema-driven \ac{LLM} extraction was explored (zero-shot or with limited supervision), but these approaches exhibited recurring practical issues such as span misalignment and hallucinated markup that required non-trivial post-processing and, in most cases, resulted in under-performing tranformer-based systems.

For \ac{M-RE}, most approaches framed the task as classification over candidate entity pairs, using explicit entity markers (or query-style formulations) to condition predictions and mitigate the large negative space via negative sampling and class rebalancing. Performance improvements were most often associated with stronger candidate construction and sampling strategies, dataset cleaning (e.g., sampling or filtering extreme relation density cases), and ensembling or fusion schemes. A few submissions explored alternative formulations (e.g., document-level interaction modeling, end-to-end relation generation, or multi-stage reasoning with auxiliary \ac{LLM} components), but the general trend remained that fully prompting-based, zero-shot \ac{LLM} solutions were not competitive on these fine-grained biomedical \ac{RE} settings and tended to be highly unreliable.

For a complete, method-by-method discussion, implementation details, and the full set of results for all submitted runs, we refer the reader to the shared task overview \cite{martinelli2025overview}.

\begin{table*}[!htb]
\centering
\caption{Performance metrics of each team’s top run for NER. For each evaluation metric, the best result is in bold, the second-best is underlined. Runs are ranked based on micro-averaged $F_1$-score}
\label{tab:NER_leaderboard}
\resizebox{\textwidth}{!}{%
\begin{tabular}{lllcccccc}
 &
   &
   &
  \multicolumn{3}{c}{\cellcolor[HTML]{EAD1DC}\textbf{Macro-avg}} &
  \multicolumn{3}{c}{\cellcolor[HTML]{EAD1DC}\textbf{Micro-avg}} \\
\rowcolor[HTML]{BDBDBD} 
\textbf{Team ID} &
  \textbf{Country} &
  \textbf{Affiliation} &
  \textbf{$P$} &
  \textbf{$R$} &
  \textbf{$F_1$} &
  \textbf{$P$} &
  \textbf{$R$} &
  \textbf{$F_1$} \\
\cellcolor[HTML]{FFFFFF}GutUZH &
  \cellcolor[HTML]{FFFFFF}Switzerland &
  \cellcolor[HTML]{FFFFFF}University of Zurich &
  \cellcolor[HTML]{83C380}0.7950 &
  \cellcolor[HTML]{96C67C}0.7736 &
  \cellcolor[HTML]{57BB8A}\textbf{0.7613} &
  \cellcolor[HTML]{57BB8A}\textbf{0.8384} &
  \cellcolor[HTML]{86C380}0.8432 &
  \cellcolor[HTML]{57BB8A}\textbf{0.8408} \\
\cellcolor[HTML]{F3F3F3}Gut-Instincts &
  \cellcolor[HTML]{F3F3F3}Denmark &
  \cellcolor[HTML]{F3F3F3}Aalborg University &
  \cellcolor[HTML]{BACB74}0.7619 &
  \cellcolor[HTML]{7AC182}0.7813 &
  \cellcolor[HTML]{5FBD88}{\ul 0.7591} &
  \cellcolor[HTML]{7CC182}0.8286 &
  \cellcolor[HTML]{6ABE86}0.8480 &
  \cellcolor[HTML]{61BD87}{\ul 0.8382} \\
\cellcolor[HTML]{FFFFFF}NLPatVCU &
  \cellcolor[HTML]{FFFFFF}United States &
  \cellcolor[HTML]{FFFFFF}Virginia Commonwealth University &
  \cellcolor[HTML]{64BE87}{\ul 0.8139} &
  \cellcolor[HTML]{FDD267}0.7161 &
  \cellcolor[HTML]{EBD36A}0.7169 &
  \cellcolor[HTML]{88C37F}0.8255 &
  \cellcolor[HTML]{65BE87}{\ul 0.8488} &
  \cellcolor[HTML]{66BE86}0.8370 \\
\cellcolor[HTML]{F3F3F3}ICUE &
  \cellcolor[HTML]{F3F3F3}United Kingdom &
  \cellcolor[HTML]{F3F3F3}The University of Edinburgh &
  \cellcolor[HTML]{57BB8A}\textbf{0.8216} &
  \cellcolor[HTML]{FDD666}0.7451 &
  \cellcolor[HTML]{6EBF85}0.7546 &
  \cellcolor[HTML]{5DBC88}{\ul 0.8369} &
  \cellcolor[HTML]{D4CF6F}0.8294 &
  \cellcolor[HTML]{75C083}0.8331 \\
\cellcolor[HTML]{FFFFFF}LYX-DMIIP-FDU &
  \cellcolor[HTML]{FFFFFF}China &
  \cellcolor[HTML]{FFFFFF}Fudan University &
  \cellcolor[HTML]{BCCC74}0.7605 &
  \cellcolor[HTML]{57BB8A}\textbf{0.7910} &
  \cellcolor[HTML]{B0CA77}0.7347 &
  \cellcolor[HTML]{E0D16C}0.8020 &
  \cellcolor[HTML]{57BB8A}\textbf{0.8513} &
  \cellcolor[HTML]{90C57D}0.8259 \\
\cellcolor[HTML]{F3F3F3}ata2425ds &
  \cellcolor[HTML]{F3F3F3}Italy &
  \cellcolor[HTML]{F3F3F3}University of Padua &
  \cellcolor[HTML]{FFD666}0.7199 &
  \cellcolor[HTML]{DAD16D}0.7546 &
  \cellcolor[HTML]{DBD16D}0.7217 &
  \cellcolor[HTML]{FED567}0.7914 &
  \cellcolor[HTML]{86C380}0.8432 &
  \cellcolor[HTML]{60BD88}0.8164 \\
\cellcolor[HTML]{FFFFFF}greenday &
  \cellcolor[HTML]{FFFFFF}United States &
  \cellcolor[HTML]{FFFFFF}Stony Brook University &
  \cellcolor[HTML]{E4D26B}0.7368 &
  \cellcolor[HTML]{AAC978}0.7682 &
  \cellcolor[HTML]{87C37F}0.7471 &
  \cellcolor[HTML]{F7D567}0.7956 &
  \cellcolor[HTML]{DDD16D}0.8278 &
  \cellcolor[HTML]{C8CE71}0.8114 \\
\cellcolor[HTML]{F3F3F3}Graphswise-1 &
  \cellcolor[HTML]{F3F3F3}Bulgaria &
  \cellcolor[HTML]{F3F3F3}Graphwise &
  \cellcolor[HTML]{AEC977}0.7691 &
  \cellcolor[HTML]{FED567}0.7398 &
  \cellcolor[HTML]{E6D26B}0.7185 &
  \cellcolor[HTML]{CFCF70}0.8066 &
  \cellcolor[HTML]{FDD267}0.7955 &
  \cellcolor[HTML]{F0D469}0.8010 \\
\cellcolor[HTML]{C9DAF8}BASELINE &
  \cellcolor[HTML]{C9DAF8}-- &
  \cellcolor[HTML]{C9DAF8}-- &
  \cellcolor[HTML]{FDD067}0.6883 &
  \cellcolor[HTML]{A7C879}0.7690 &
  \cellcolor[HTML]{FED567}0.7047 &
  \cellcolor[HTML]{FDD067}0.7639 &
  \cellcolor[HTML]{F4D568}0.8238 &
  \cellcolor[HTML]{FED567}0.7927 \\
\cellcolor[HTML]{F3F3F3}ataupd2425-gainer &
  \cellcolor[HTML]{F3F3F3}Italy &
  \cellcolor[HTML]{F3F3F3}University of Padua &
  \cellcolor[HTML]{F8BC6A}0.5808 &
  \cellcolor[HTML]{F6B86B}0.5322 &
  \cellcolor[HTML]{F7BB6A}0.5281 &
  \cellcolor[HTML]{6BBF85}0.8333 &
  \cellcolor[HTML]{FBCA68}0.7397 &
  \cellcolor[HTML]{FED367}0.7837 \\
\cellcolor[HTML]{FFFFFF}DS@GT-bioasq-task6 &
  \cellcolor[HTML]{FFFFFF}United States &
  \cellcolor[HTML]{FFFFFF}NA &
  \cellcolor[HTML]{FAC669}0.6342 &
  \cellcolor[HTML]{6EBF85}{\ul 0.7849} &
  \cellcolor[HTML]{FED267}0.6872 &
  \cellcolor[HTML]{FBC968}0.7337 &
  \cellcolor[HTML]{FED567}0.8197 &
  \cellcolor[HTML]{FDD267}0.7743 \\
\cellcolor[HTML]{F3F3F3}DS@GT-BioNER &
  \cellcolor[HTML]{F3F3F3}Canada &
  \cellcolor[HTML]{F3F3F3}NA &
  \cellcolor[HTML]{FCCD68}0.6731 &
  \cellcolor[HTML]{FBC868}0.6497 &
  \cellcolor[HTML]{FCCC68}0.6469 &
  \cellcolor[HTML]{FED267}0.7783 &
  \cellcolor[HTML]{FBCA68}0.7437 &
  \cellcolor[HTML]{FDD067}0.7606 \\
\cellcolor[HTML]{FFFFFF}ataupd2425-pam &
  \cellcolor[HTML]{FFFFFF}Italy &
  \cellcolor[HTML]{FFFFFF}University of Padua &
  \cellcolor[HTML]{FBC769}0.6400 &
  \cellcolor[HTML]{FED567}0.7435 &
  \cellcolor[HTML]{FDD167}0.6763 &
  \cellcolor[HTML]{F8BF6A}0.6809 &
  \cellcolor[HTML]{FDCF67}0.7745 &
  \cellcolor[HTML]{FBCA68}0.7247 \\
\cellcolor[HTML]{F3F3F3}Schemalink &
  \cellcolor[HTML]{F3F3F3}Italy &
  \cellcolor[HTML]{F3F3F3}University of Milan &
  \cellcolor[HTML]{F3AA6D}0.4813 &
  \cellcolor[HTML]{F5B46B}0.5038 &
  \cellcolor[HTML]{F5B26C}0.4650 &
  \cellcolor[HTML]{F1A56D}0.5547 &
  \cellcolor[HTML]{F4B16C}0.5659 &
  \cellcolor[HTML]{F4AE6C}0.5602 \\
\cellcolor[HTML]{FFFFFF}BIU-ONLP &
  \cellcolor[HTML]{FFFFFF}Israel &
  \cellcolor[HTML]{FFFFFF}Bar Ilan University &
  \cellcolor[HTML]{F0A36E}0.4393 &
  \cellcolor[HTML]{EF9F6E}0.3585 &
  \cellcolor[HTML]{F1A56E}0.3711 &
  \cellcolor[HTML]{EE996F}0.4916 &
  \cellcolor[HTML]{F1A36E}0.4721 &
  \cellcolor[HTML]{F0A26E}0.4816 \\
\rowcolor[HTML]{E67C73} 
\cellcolor[HTML]{F3F3F3}lasigeBioTM &
  \cellcolor[HTML]{F3F3F3}Portugal &
  \cellcolor[HTML]{F3F3F3}Universidade de Lisboa &
  0.2206 &
  0.1034 &
  0.0863 &
  0.3471 &
  0.1964 &
  0.2509
\end{tabular}%
}
\end{table*}

\begin{table*}[!htb]
\centering
\caption{Performance metrics of each team’s top run for M-RE. For each evaluation metric, the best result is in bold, the second-best is underlined. Runs are ranked based on micro-averaged $F_1$-score}
\label{tab:TM-RE_leaderboard}
\resizebox{\textwidth}{!}{%
\begin{tabular}{lllcccccc}
 &
   &
   &
  \multicolumn{3}{c}{\cellcolor[HTML]{EAD1DC}\textbf{Macro-avg}} &
  \multicolumn{3}{c}{\cellcolor[HTML]{EAD1DC}\textbf{Micro-avg}} \\
\rowcolor[HTML]{BDBDBD} 
\textbf{Team ID} &
  \textbf{Country} &
  \textbf{Affiliation} &
  \textbf{$P$} &
  \textbf{$R$} &
  \textbf{$F_1$} &
  \textbf{$P$} &
  \textbf{$R$} &
  \textbf{$F_1$} \\
\cellcolor[HTML]{FFFFFF}Gut-Instincts &
  \cellcolor[HTML]{FFFFFF}Denmark &
  \cellcolor[HTML]{FFFFFF}Aalborg University &
  \cellcolor[HTML]{72C084}0.3310 &
  \cellcolor[HTML]{86C380}{\ul 0.4303} &
  \cellcolor[HTML]{57BB8A}\textbf{0.3497} &
  \cellcolor[HTML]{94C57C}0.4215 &
  \cellcolor[HTML]{8BC47E}{\ul 0.5147} &
  \cellcolor[HTML]{57BB8A}\textbf{0.4635} \\
\cellcolor[HTML]{F3F3F3}Graphswise-1 &
  \cellcolor[HTML]{F3F3F3}Bulgaria &
  \cellcolor[HTML]{F3F3F3}Graphwise &
  \cellcolor[HTML]{70BF84}{\ul 0.3323} &
  \cellcolor[HTML]{FFD666}0.2369 &
  \cellcolor[HTML]{C5CD72}0.2603 &
  \cellcolor[HTML]{6FBF84}{\ul 0.4686} &
  \cellcolor[HTML]{FDD267}0.3097 &
  \cellcolor[HTML]{A8C878}{\ul 0.3729} \\
\cellcolor[HTML]{FFFFFF}ICUE &
  \cellcolor[HTML]{FFFFFF}United Kingdom &
  \cellcolor[HTML]{FFFFFF}The University of Edinburgh &
  \cellcolor[HTML]{D8D06E}0.2509 &
  \cellcolor[HTML]{8AC47F}0.4239 &
  \cellcolor[HTML]{AAC978}{\ul 0.2825} &
  \cellcolor[HTML]{FFD666}0.2858 &
  \cellcolor[HTML]{91C57D}0.5054 &
  \cellcolor[HTML]{AFCA77}0.3651 \\
\cellcolor[HTML]{F3F3F3}LYX-DMIIP-FDU &
  \cellcolor[HTML]{F3F3F3}China &
  \cellcolor[HTML]{F3F3F3}Fudan University &
  \cellcolor[HTML]{FDD267}0.2106 &
  \cellcolor[HTML]{FCD666}0.2418 &
  \cellcolor[HTML]{FDD067}0.1990 &
  \cellcolor[HTML]{BECC73}0.3682 &
  \cellcolor[HTML]{FED666}0.3257 &
  \cellcolor[HTML]{C0CC73}0.3457 \\
\cellcolor[HTML]{FFFFFF}ONTUG &
  \cellcolor[HTML]{FFFFFF}Austria &
  \cellcolor[HTML]{FFFFFF}University of Graz + Ontotext &
  \cellcolor[HTML]{CECF70}0.2589 &
  \cellcolor[HTML]{FED367}0.2293 &
  \cellcolor[HTML]{EED469}0.2266 &
  \cellcolor[HTML]{CBCE71}0.3529 &
  \cellcolor[HTML]{FFD666}0.3231 &
  \cellcolor[HTML]{C8CE71}0.3373 \\
\cellcolor[HTML]{C9DAF8}BASELINE &
  \cellcolor[HTML]{C9DAF8}-- &
  \cellcolor[HTML]{C9DAF8}-- &
  \cellcolor[HTML]{57BB8A}\textbf{0.3514} &
  \cellcolor[HTML]{F9C169}0.1829 &
  \cellcolor[HTML]{FFD666}0.2123 &
  \cellcolor[HTML]{57BB8A}\textbf{0.4986} &
  \cellcolor[HTML]{F8C06A}0.2453 &
  \cellcolor[HTML]{CFCF70}0.3288 \\
\cellcolor[HTML]{FFFFFF}Schemalink &
  \cellcolor[HTML]{FFFFFF}Italy &
  \cellcolor[HTML]{FFFFFF}University of Milan &
  \cellcolor[HTML]{F7D567}0.2265 &
  \cellcolor[HTML]{93C57D}0.4088 &
  \cellcolor[HTML]{CCCE71}0.2546 &
  \cellcolor[HTML]{F7B96B}0.1948 &
  \cellcolor[HTML]{A9C978}0.4665 &
  \cellcolor[HTML]{FFD666}0.2749 \\
\cellcolor[HTML]{F3F3F3}ataupd2425-pam &
  \cellcolor[HTML]{F3F3F3}Italy &
  \cellcolor[HTML]{F3F3F3}University of Padua &
  \cellcolor[HTML]{FCCB68}0.1940 &
  \cellcolor[HTML]{E7D36B}0.2764 &
  \cellcolor[HTML]{FDD067}0.1982 &
  \cellcolor[HTML]{F9C369}0.2278 &
  \cellcolor[HTML]{F3D568}0.3432 &
  \cellcolor[HTML]{FED567}0.2738 \\
\cellcolor[HTML]{FFFFFF}ataupd2425-gainer &
  \cellcolor[HTML]{FFFFFF}Italy &
  \cellcolor[HTML]{FFFFFF}University of Padua &
  \cellcolor[HTML]{FFD666}0.2203 &
  \cellcolor[HTML]{F4B06C}0.1384 &
  \cellcolor[HTML]{F8BD6A}0.1538 &
  \cellcolor[HTML]{90C57D}0.4272 &
  \cellcolor[HTML]{F4AE6C}0.1810 &
  \cellcolor[HTML]{FDCF67}0.2542 \\
\cellcolor[HTML]{F3F3F3}NLPatVCU &
  \cellcolor[HTML]{F3F3F3}United States &
  \cellcolor[HTML]{F3F3F3}Virginia Commonwealth University &
  \cellcolor[HTML]{F7BA6B}0.1522 &
  \cellcolor[HTML]{57BB8A}\textbf{0.5041} &
  \cellcolor[HTML]{FBD667}0.2163 &
  \cellcolor[HTML]{F2A86D}0.1423 &
  \cellcolor[HTML]{57BB8A}\textbf{0.6005} &
  \cellcolor[HTML]{FAC769}0.2300 \\
\cellcolor[HTML]{FFFFFF}BIU-ONLP &
  \cellcolor[HTML]{FFFFFF}Israel &
  \cellcolor[HTML]{FFFFFF}Bar Ilan University &
  \cellcolor[HTML]{F3AB6D}0.1171 &
  \cellcolor[HTML]{EF9C6F}0.0854 &
  \cellcolor[HTML]{F0A16E}0.0879 &
  \cellcolor[HTML]{FAC569}0.2339 &
  \cellcolor[HTML]{F1A46E}0.1461 &
  \cellcolor[HTML]{F6B66B}0.1799 \\
\cellcolor[HTML]{F3F3F3}ToGS &
  \cellcolor[HTML]{F3F3F3}Austria &
  \cellcolor[HTML]{F3F3F3}University of Graz &
  \cellcolor[HTML]{E88672}0.0249 &
  \cellcolor[HTML]{E78273}0.0180 &
  \cellcolor[HTML]{E88472}0.0203 &
  \cellcolor[HTML]{F4B16C}0.1702 &
  \cellcolor[HTML]{EA8A71}0.0536 &
  \cellcolor[HTML]{ED9670}0.0815 \\
\rowcolor[HTML]{E67C73} 
\cellcolor[HTML]{FFFFFF}lasigeBioTM &
  \cellcolor[HTML]{FFFFFF}Portugal &
  \cellcolor[HTML]{FFFFFF}Universidade de Lisboa &
  0.0000 &
  0.0000 &
  0.0000 &
  0.0000 &
  0.0000 &
  0.0000
\end{tabular}%
}
\end{table*}

\section{Human Performance}
To estimate human-level performance on the benchmark tasks proposed in the evaluation campaign, we evaluated layperson annotations on the shared honeypot documents.
Each student's annotations were treated as an individual system submission and evaluated using the same script and metrics applied to participant test runs, with the final annotated version of each honeypot document used as ground truth.
To establish a fair comparison, we re-trained our baseline system leaving out the honeypot documents to prevent data leakage, then ran inference on the honeypot set and evaluated its predictions.

For the \ac{NER} task, all laypeople achieved micro-averaged precision, recall, and $F_1$ scores above 0.40, with average scores of 0.79 precision, 0.77 recall, and 0.77 $F_1$. 
Although lower, results were still robust for \ac{M-RE}, where, on average, laypeople scored higher on precision (0.61) and slightly lower on recall (0.52) and consequently on $F_1$ (0.53). 
The baseline system achieved a micro-averaged precision, recall, and $F_1$ of 0.83 for \ac{NER}, and 0.44 precision, 0.31 recall, and 0.37 $F_1$ for \ac{M-RE}.
These results indicate that, while \ac{NER} can be effectively tackled by automatic systems, achieving results comparable to those of non-expert annotators, \ac{RE} remains significantly more complex. Indeed, in this task layperson annotators consistently outperformed the baseline across all metrics, highlighting the semantic and contextual difficulty of our benchmark.

\section{Conceptual Schema}
\begin{figure*}
    \centering
    \includegraphics[width=0.9\linewidth]{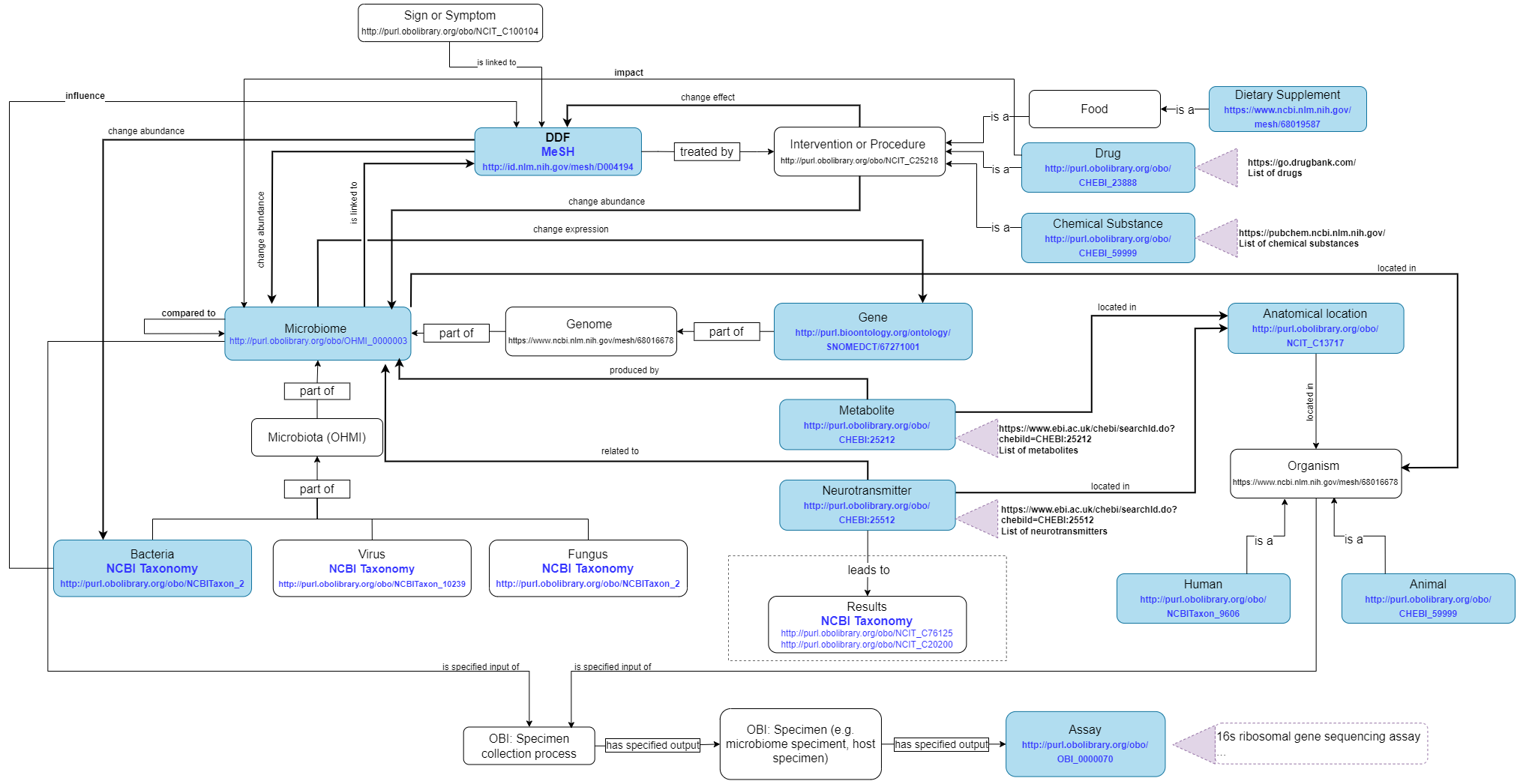}
    \caption{Conceptual schema defining entity types and relation predicates captured within \textsc{GutBrainIE}. Blue rectangles represent annotated entity types, while white rectangles indicate concepts considered during schema design but excluded from annotation due to low frequency. Arrows indicate valid relation directions and predicates between entities.}
    \label{fig:conceptual_schema}
\end{figure*}

Entities and relations to be annotated, and thus to be predicted by \ac{IE} systems, within the \textsc{GutBrainIE} benchmark are defined by the conceptual schema shown in Figure \ref{fig:conceptual_schema}. It was collaboratively designed by expert annotators and subsequently validated by external biomedical specialists.
During the initial development phase, we explored a wider range of entity types and relation predicates. However, after preliminary pilot annotations, we filtered out those that were excessively underrepresented. 

\section{Types of Named Entities}
\textsc{GutBrainIE} includes annotations for 13 distinct entity types, listed in Table \ref{tab:entity_definitions}. Each entity type is associated with a unique \ac{URI} that links it to a standardized concept in a reference vocabulary and is accompanied by an explanation that defines its scope and semantic meaning.
Moreover, Figure \ref{fig:NER_Ents_Distribution} shows the distribution of entity types across the full \textsc{GutBrainIE} dataset and within the manually and automatically annotated subsets.  

\begin{table*}[!htb]
    \caption{Overview of the 13 entity labels used in the \textsc{GutBrainIE} corpus, including their corresponding \acp{URI} and explanations.}
    \label{tab:entity_definitions}
    \centering
    \resizebox{0.75\textwidth}{!}{%
    \begin{tabular}{L{0.2\linewidth}M{0.25\linewidth}L{0.45\linewidth}}
        \hline
        \textbf{Entity Label} & \textbf{URI} & \textbf{Explanation} \\ \hline
        Anatomical Location & NCIT\_C13717 & Named locations of or within the body. \\ \hline
        Animal & NCIT\_C14182 & A non-human living organism that has membranous cell walls, requires oxygen and organic foods, and is capable of voluntary movement, as distinguished from a plant or mineral. \\ \hline
        Biomedical Technique & NCIT\_C15188 & Research concerned with the application of biological and physiological principles to clinical medicine. \\ \hline
        Bacteria & NCBITaxon\_2 & One of the three domains of life (the others being Eukarya and ARCHAEA), also called Eubacteria. They are unicellular prokaryotic microorganisms which generally possess rigid cell walls, multiply by cell division, and exhibit three principal forms: round or coccal, rodlike or bacillary, and spiral or spirochetal. \\ \hline
        Chemical & CHEBI\_59999 & A chemical substance is a portion of matter of constant composition, composed of molecular entities of the same type or of different types. This category also includes metabolites, which in biochemistry are the intermediate or end product of metabolism, and neurotransmitters, which are endogenous compounds used to transmit information across the synapses. \\ \hline
        Dietary Supplement & MESH\_68019587 & Products in capsule, tablet or liquid form that provide dietary ingredients, and that are intended to be taken by mouth to increase the intake of nutrients. Dietary supplements can include macronutrients, such as proteins, carbohydrates, and fats; and/or micronutrients, such as vitamins; minerals; and phytochemicals. \\ \hline
        Disease, Disorder, or Finding (DDF) & NCIT\_C7057 & A condition that is relevant to human neoplasms and non-neoplastic disorders. This includes observations, test results, history and other concepts relevant to the characterization of human pathologic conditions. \\ \hline
        Drug & CHEBI\_23888 & Any substance which when absorbed into a living organism may modify one or more of its functions. The term is generally accepted for a substance taken for a therapeutic purpose, but is also commonly used for abused substances. \\ \hline
        Food & NCIT\_C1949 & A substance consumed by humans and animals for nutritional purpose. \\ \hline
        Gene & SNOMEDCT\_67261001 & A functional unit of heredity which occupies a specific position on a particular chromosome and serves as the template for a product that contributes to a phenotype or a biological function. \\ \hline
        Human & NCBITaxon\_9606 & Members of the species Homo sapiens. \\ \hline
        Microbiome & OHMI\_0000003 & This term refers to the entire habitat, including the microorganisms (bacteria, archaea, lower and higher eukaryotes, and viruses), their genomes (i.e., genes), and the surrounding environmental conditions. \\ \hline
        Statistical Technique & NCIT\_C19044 & A method of calculating, analyzing, or representing statistical data. \\ \hline
    \end{tabular}
    }
\end{table*}

\section{Types of Relations}
\begin{figure*}[!htb]
    \centering
    \begin{subfigure}[b]{0.45\linewidth}
        \centering
        \includegraphics[width=\linewidth]{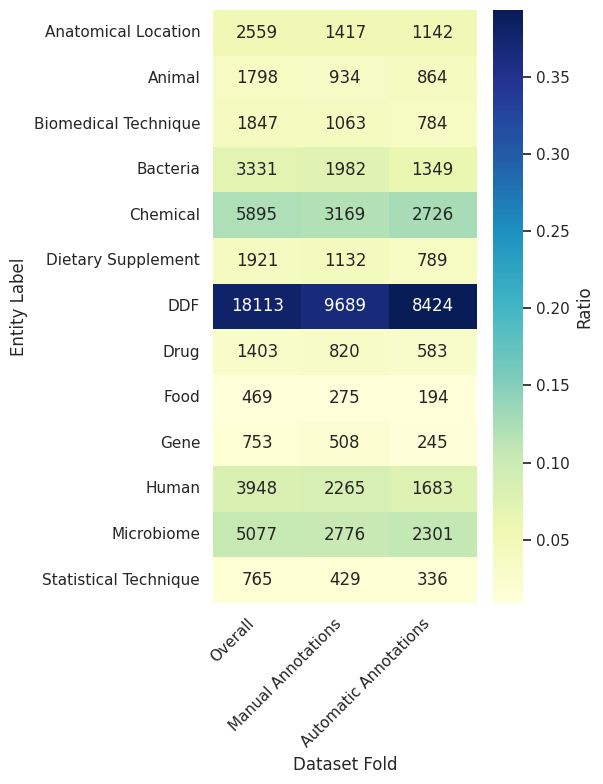}
        \caption{Distribution of annotated entity mentions across dataset folds and entity labels.}
        \label{fig:NER_Ents_Distribution}
    \end{subfigure}
    \hfill
    \begin{subfigure}[b]{0.45\linewidth}
        \centering
        \includegraphics[width=\linewidth]{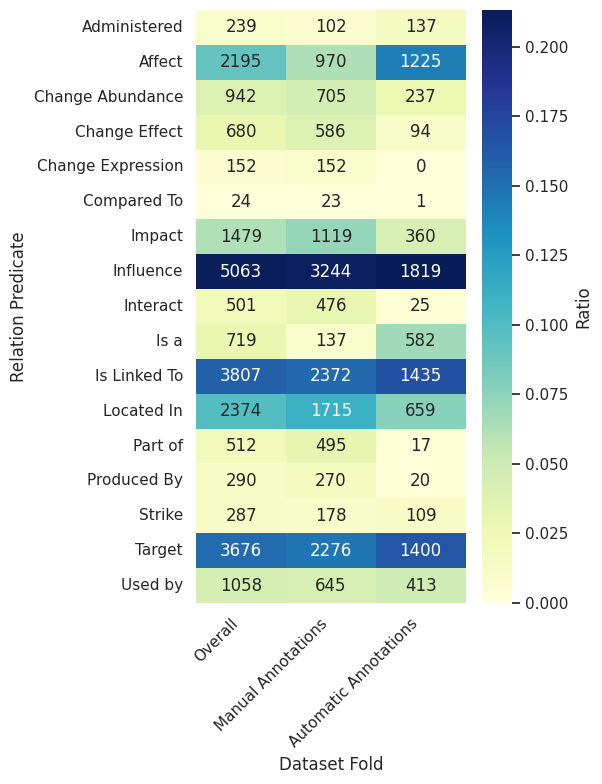}
        \caption{Distribution of annotated relations across dataset folds and relation predicates.}
        \label{fig:RE_Rels_Distribution}
    \end{subfigure}
    \caption{Distribution of entity types and relation predicates across manually and automatically annotated folds.}
    \label{fig:NER_RE_Distributions}
\end{figure*}

\textsc{GutBrainIE} features annotations for 17 distinct relation predicates, each of which possibly connects multiple combinations of head and tail entity types, resulting in over 50 possible (\textit{head, predicate, tail}) triples. 
Table \ref{tab:relation_definitions} lists all relation predicates used in \textsc{GutBrainIE}, represented as (\textit{head, predicate, tail}) combinations according to the conceptual schema depicted in Figure \ref{fig:conceptual_schema}.
In addition, Figure \ref{fig:RE_Rels_Distribution} illustrates the distribution of relation predicates across the full \textsc{GutBrainIE} dataset and within the manually and automatically annotated subsets.

\begin{table*}[!htb]
    \caption{Overview of the relations used in the \textsc{GutBrainIE} corpus, expressed as head-predicate-tail triples.}    
    \label{tab:relation_definitions}
    \centering
    \resizebox{0.7\textwidth}{!}{%
    \begin{tabular}{L{0.3\linewidth}L{0.3\linewidth}L{0.2\linewidth}}
        \hline
        \textbf{Head Entity} & \textbf{Tail Entity} & \textbf{Predicate} \\
        \hline
        Anatomical Location & Human \newline Animal & (1) Located in \\ \hline
        Bacteria & Bacteria \newline Chemical \newline Drug & (2) Interact \\ \hline
        Bacteria & DDF & (3) Influence \\ \hline
        Bacteria & Gene & (4) Change expression \\ \hline
        Bacteria & Human \newline Animal & (1) Located in \\ \hline
        Bacteria & Microbiome & (5) Part of \\ \hline
        Chemical & Anatomical Location \newline Human \newline Animal & (1) Located in \\ \hline
        Chemical & Chemical & (2) Interact \newline (5) Part of \\ \hline
        Chemical & Microbiome & (6) Impact \newline (7) Produced by \\ \hline
        Chemical \newline Dietary Supplement \newline Drug \newline Food & Bacteria \newline Microbiome & (6) Impact \\ \hline
        Chemical \newline Dietary Supplement \newline Food & DDF & (3) Influence \\ \hline
        Chemical \newline Dietary Supplement \newline Drug \newline Food & Gene & (4) Change expression \\ \hline
        Chemical \newline Dietary Supplement \newline Drug \newline Food & Human \newline Animal & (8) Administered \\ \hline
        DDF & Anatomical Location & (9) Strike \\ \hline
        DDF & Bacteria \newline Microbiome & (10) Change abundance \\ \hline
        DDF & Chemical & (2) Interact \\ \hline
        DDF & DDF & (11) Affect \newline (12) Is a \\ \hline
        DDF & Human \newline Animal & (13) Target \\ \hline
        Drug & Chemical \newline Drug & (2) Interact \\ \hline
        Drug & DDF & (14) Change effect \\ \hline
        Human \newline Animal \newline Microbiome & Biomedical Technique & (15) Used by \\ \hline
        Microbiome & Anatomical Location \newline Human \newline Animal & (1) Located in \\ \hline
        Microbiome & Gene & (4) Change expression \\ \hline
        Microbiome & DDF & (16) Is linked to \\ \hline
        Microbiome & Microbiome & (17) Compared to \\
        \hline
    \end{tabular}
    }
\end{table*}

\section{Types of Concept-Level Links}
To support semantic normalization and concept-level reasoning, entities annotated in the Platinum and Gold collections have been linked to concepts in standardized biomedical vocabularies. We tried our best to minimize the number of different vocabularies employed, resulting in a total of six biomedical vocabularies and a custom-defined ontology. Each entity type is linked to a number of vocabularies ranging from 3 to 6. The vocabularies employed for each entity type are reported in Table \ref{tab:links_definitions}, which also includes, for each vocabulary and entity type, the number of different unique \acp{URI} employed. 

\begin{table*}[!htb]
\centering
\caption{Biomedical vocabularies used for concept-level linking of entity mentions. For each entity label, the table lists the reference vocabularies, ordered accordingly to the priority considered in the \ac{NEL} pipeline (see Section \ref{sec:DataCollection}), and reports for each of these the number of unique \acp{URI} assigned. \textit{GBIE} indicates our custom-defined ontology.}
\label{tab:links_definitions}
\resizebox{0.85\textwidth}{!}{%
\begin{tabular}{ll}
\hline
\textbf{Entity Label} & \textbf{Linked Vocabularies}                                        \\ \hline
Anatomical Location   & UMLS (17), NCIT (70), GBIE (3)                                      \\ \hline
Animal                & UMLS (8), NCIT (7), NCBITaxon (5), GBIE (3)                         \\ \hline
Bacteria              & UMLS (23),  NCIT (6), NCBITaxon (136), MESH (34), OMIT (1), GBIE (6) \\ \hline
Biomedical Technique \qquad\qquad & UMLS (65), NCIT (16), OMIT (2), NCBITaxon (4), GBIE (53)            \\ \hline
Chemical              & UMLS (75), NCIT (94), CHEBI (209), OMIT (2), GBIE (14)              \\ \hline
Dietary Supplement & NCIT (34), UMLS (11), CHEBI (13), NCBITaxon (4), OMIT (2), MESH (2), GBIE (3) \\ \hline
DDF                   & UMLS (179), NCIT (259), OMIT (36), NCBITaxon (1), GBIE (27)         \\ \hline
Drug                  &  UMLS (22), NCIT (9), CHEBI (34), OMIT (1), NCBITaxon (1), GBIE (3)  \\ \hline
Food                  & UMLS (23), NCIT (17), GBIE (3)                                      \\ \hline
Gene                  & UMLS (52), NCIT (42), OMIT (4), CHEBI (1), GBIE (13)                \\ \hline
Human                 & UMLS (38), MESH (8), GBIE (24)                                      \\ \hline
Microbiome            & UMLS (2), NCIT (1), NCBITaxon (3), GBIE (15)                         \\ \hline
Statistical Technique & UMLS (35), NCIT (21), GBIE (23)                                     \\ \hline
\end{tabular}%
}
\end{table*}

\section{\ac{LLM}-Based Definition Generation for \ac{NEL}}
As stated in Section \ref{sec:DataCollection}, the final stage of our \ac{NEL} pipeline was reached when no match could be found between an entity mention and the defined reference vocabularies. In these cases, we created a new individual in our custom ontology and prompted a \ac{LLM} to generate an appropriate definition.
Figure \ref{fig:llm_prompt_example} shows an example of the prompt and response for the entity mention ``\textit{psychobiotics}'', labeled as a ``\textit{dietary supplement}''

\begin{figure*}[!htb]
\centering
\begin{quote}
\small
\textbf{Prompt:}

\texttt{\#\#Role\#\#} \\
\texttt{You are a biomedical expert tasked with enriching a biomedical ontology. Given the sentence below, identify the entity mention marked between **...**, which has been labeled as a dietary supplement. Write a concise and ontologically appropriate definition for this entity, suitable for use as an \texttt{rdfs:comment} or \texttt{skos:definition} in a formal biomedical knowledge base.}

\vspace{0.5em}
\texttt{\#\#Text\#\#} \\
\texttt{Psychobiotics exhibit positive effects on the intestinal barrier, immune response, cortisol levels, and the hypothalamic-pituitary-adrenal (HPA) axis. They are a new class of probiotics that have been shown to improve mental health, including the treatment [...]}

\vspace{0.5em}
\texttt{\#\#Instructions\#\#} \\
\texttt{- Output only the definition; do not include any additional commentary, introductions, or explanations.} \\
\texttt{- Define the entity as a class-level concept (not as an individual instance).} \\
\texttt{- Ensure the definition is concise (2–3 sentences), scientifically sound, and suitable for ontological use.} \\
\texttt{- Emphasize biomedical function, mechanisms of action, and therapeutic relevance.} \\
\texttt{- Use a formal, precise tone appropriate for ontology documentation.}

\vspace{0.8em}
\textbf{Generated Answer:}

\texttt{Psychobiotics are live microorganisms that, when administered in adequate amounts, produce a health benefit in the host by modulating the gut-brain axis and influencing brain function or behavior.}
\end{quote}
\caption{Example prompt and response for the LLM-based definition generation stage of our \ac{NEL} pipeline (see Section \ref{sec:DataCollection}).}
\label{fig:llm_prompt_example}
\end{figure*}

\end{document}